\documentclass[runningheads]{llncs}
\usepackage{graphicx}
\usepackage{amsmath,amssymb} 
\usepackage{color}

\renewcommand{\vec}[1]{\mathbf{#1}}
\DeclareMathOperator*{\argminA}{arg\,min} 
\newcommand{\euclidean}[1]{\left\lVert #1 \right\rVert}

\begin{document}
	\title{NRST: Non-rigid Surface Tracking from Monocular Video} 
	
	\titlerunning{NRST: Non-rigid Surface Tracking from Monocular Video}
	%
	\author{Marc Habermann\inst{1}\orcidID{0000-0003-3899-7515} \and
		Weipeng Xu\inst{1}\orcidID{0000-0001-9548-5108} \and
		Helge Rhodin\inst{2}\orcidID{0000-0003-2692-0801} \and
		Michael Zollh\"ofer\inst{3}\orcidID{0000-0003-1219-0625} \and
		Gerard Pons-Moll\inst{1}\orcidID{0000-0001-5115-7794} \and \\
		Christian Theobalt\inst{1}\orcidID{0000-0001-6104-6625}}
	%
	\authorrunning{M. Habermann et al.}
	%
	%
	\institute{
		Max Planck Institute for Informatics, Saarbruecken 66123, Germany \\
		\url{https://www.mpi-inf.mpg.de/home/} \\
		\email{\{mhaberma, wxu, gpons, theobalt\}@mpi-inf.mpg.de} \and
		EPFL, Lausanne CH-1015, Switzerland\\
		\url{https://www.epfl.ch/}\\
		\email{helge.rhodin@epfl.ch} \and
		Stanford University, Stanford CA 94305, USA\\
		\url{https://www.stanford.edu/} \\
		\email{zollhoefer@cs.stanford.edu}
	}
	\maketitle              
	\begin{abstract}
We propose an efficient method for non-rigid surface tracking from monocular RGB videos.
Given a video and a template mesh, our algorithm sequentially registers the template non-rigidly to each frame.
We formulate the per-frame registration as an optimization problem that includes a novel texture term specifically tailored towards tracking objects with uniform texture but fine-scale structure, such as the regular micro-structural patterns of fabric.
Our texture term exploits the orientation information in the micro-structures of the objects, e.g., the yarn patterns of fabrics. 
This enables us to accurately track uniformly colored materials that have these high frequency micro-structures, for which traditional photometric terms are usually less effective.
The results demonstrate the effectiveness of our method on both general textured non-rigid objects and monochromatic fabrics.
%
		%
	\end{abstract}
	\begin{figure}
		\includegraphics[width=\textwidth]{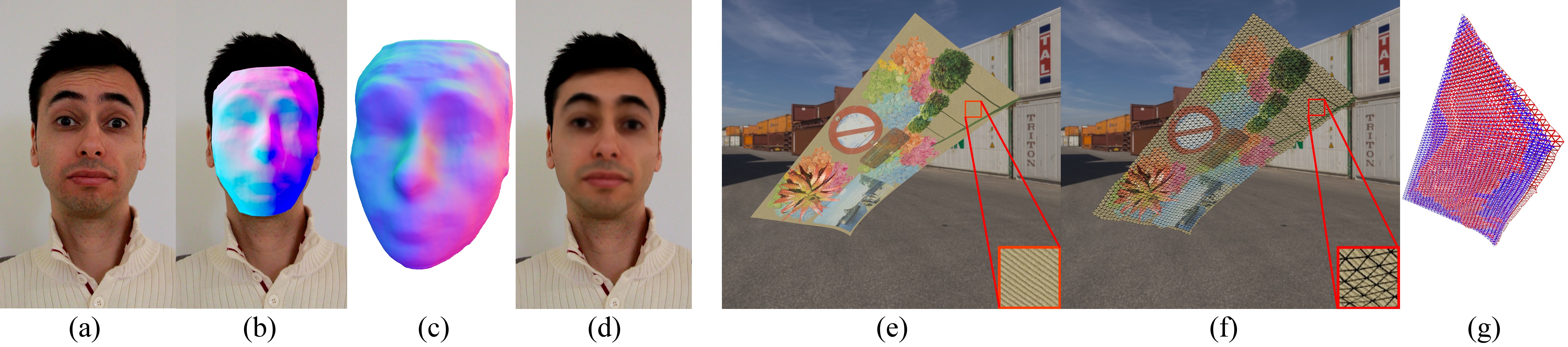}
		\caption
		{
			We propose an efficient method for interactive non-rigid surface tracking from monocular RGB videos for general objects such as faces ((a)-(d)). 
			Given the input image (a) our reconstruction nicely overlays with the input (b) and looks also plausible from another view point (c). 
			The textured overlay looks realistic as well (d).
			Furthermore, our novel texture term leads to improved reconstruction quality for fabrics given a single video (e).
			Again the overlayed reconstruction (f) aligns well, and also in 3D (g) our result (red) matches the ground truth (blue).
		}
		\label{fig:teaser}
	\end{figure}
	\section{Introduction}
\label{sec:intro}
%
%
In this paper, we propose NRST, an efficient method for non-rigid surface tracking from monocular RGB videos.
Capturing the non-rigid deformation of a dynamic surface is an important and long-standing problem in computer vision.
It has a wide range of real world applications in fields such as virtual/augmented reality, medicine and visual effects.
Most of the existing methods are based on multi-view imagery, where expensive and complicated system setups are required~\cite{car001,pon005,per007}.
There also exist methods that rely on only a single depth or RGB-D camera~\cite{xu2014eccv,zol014,new011,new015}.
However, these sensors are not as ubiquitous as RGB cameras, and these methods cannot be applied on plenty of existing video footage which is found on social media like YouTube.
There are also monocular RGB methods~\cite{yu015,sal007}, of course with their own limitations; e.g., they rely on highly textured surfaces and they are often times slow.
%
%
\par
In this work, we present a method which is able to densely track the non-rigid deformations of general objects such as faces and fabrics from a single RGB video. 
To solve this challenging problem, our method relies on a textured mesh template of the deforming object's surface.
Given the input video, our algorithm sequentially registers the template to each frame.
More specifically, our method automatically reproduces a deformation sequence of the template model that coincides with the non-rigid surface motion in the video. 
To this end, we formulate the per-frame registration as a non-linear least squares optimization problem -- with an objective function consisting of a photometric alignment and several regularization terms.
The optimization is computationally intensive due to the large number of residuals in our alignment objective.
To address this, we adapt the efficient GPU-based Gauss-Newton solver of Zollhoefer et al.~\cite{zol014} to our problem that allows for deformable object tracking at interactive frame rates.
%
%
\par
Besides the efficiency of the algorithm, the core contribution of our approach is a novel texture term that exploits the orientation information in the micro-structures of the tracked objects, such as the yarn patterns of fabrics. 
This enables us to track uniformly colored materials which have high frequency patterns, for which the classical color-based term is usually less effective.
%
%
\par
In our experimental results, we evaluate our method qualitatively and quantitatively on several challenging sequences of deforming surfaces. We use well established benchmarks, such as pieces of cloth~\cite{var012,sal008} and human faces~\cite{yu015,valg012}.
The results demonstrate that our method can accurately track general non-rigid objects.
Furthermore, for materials with regular micro-structural patterns, such as fabrics, the tracking accuracy is further improved with our texture term.

	\section{Related Work}
\label{sec:relatedwork}
%
%
There is a variety of approaches that reconstruct geometry from multiple images, e.g., template-free methods~\cite{car001}, variational ones~\cite{pon005} or object specific approaches~\cite{per007}.
Although multi-view methods can produce accurate tracking results, their setup is expensive and hard to operate. 
%
%
Some approaches use a single RGB-D sensor instead~\cite{xu2014eccv,zol014,new011,new015,jordt2011fast,jordt2013direct,DoubleFusion2018}. 
They manage to capture deformable surfaces nicely and at high efficiency, some even build up a template model alongside per-frame reconstruction.
The main limitations of these methods are that the sensors have a high power consumption, they do not work outdoors, the object has to be close to the camera and they cannot use the large amount of RGB-only video footage provided by social media.
On these grounds, we aim for a method that uses just a single RGB video as input.
In the following, we focus on related monocular reconstruction and tracking approaches.
%
%
%
%
\par
\textbf{Monocular Methods.} 
Non-rigid structure from motion methods, which do not rely on any template, try to infer the 3D geometry from a single video by using a prior-free formulation~\cite{dai014}, global models~\cite{tor008}, local ones~\cite{rus011} or solving a variational formulation~\cite{gar013}.
But they often either capture the deformations only coarsely, are not able to model strong deformations, typically require strongly textured objects or rely on dense 2D correspondences.
%
%
By constraining the setting to specific types of objects such as faces~\cite{garr013}, very accurate reconstructions can be obtained, but at the expense of generality.
%
%
Since in recent years, several approaches~\cite{pan009,lab007} build a 3D model given a set of images, and even commercial software\footnote{\url{http://www.agisoft.com/}} is available for this task, template acquisition has become easier.
%
%
Templates are an effective prior for the challenging task of estimating non-rigid deformations from single images as demonstrated by previous work~\cite{sal007,sal008,sal2008,she009,bru010,per011,mal011,oest012,bar012,var012,yu015,mor009,sal009,mal013,sal011,wen016}.
But even if a template is used, ambiguities~\cite{sal007} remain and additional constraints have to be imposed. 
Theoretical results~\cite{bar012} show that only allowing isometric deformations~\cite{per011} results in a uniquely defined solution.
Therefore, approaches constrain the deformation space in several ways, e.g., by a Laplacian regularization~\cite{oest012} or by non-linear~\cite{sal2008} or linear local surface models~\cite{sal011}.
Salzmann et al.~\cite{sal009} argued that relaxing the isometric constraint is beneficial since it allows to model sharp folds.
Moreno-Noguer et al.~\cite{mor009} and Malti et al.~\cite{mal013} even go beyond this and show results for elastic surfaces; Tsoli and Argyros~\cite{Tsoli:3DV:2016} demonstrated tracking surfaces that undergo topological changes but require a depth camera.
Other approaches investigate how to make reconstruction more robust under faster motions~\cite{she009} and occlusions~\cite{ngo015}, or try to replace the feature-based data term by a dense pixel-based one~\cite{mal011} and to find better texture descriptors~\cite{rao991,gar992,lia005}.
Brunet et al.~\cite{bru010} and Yu et al.~\cite{yu015} formulate the problem of estimating non-rigid deformations as minimizing an objective function which brings them closest to our formulation. 
In particular, we adopt the photometric, spatial and temporal terms of Yu et al.~\cite{yu015} and combine them with an isometric and acceleration constraint as well as our novel texture term.
%
%
\par
%
%
Along the line of monocular methods, we propose NRST, a template-based reconstruction framework that estimates the non-rigidly deforming geometry of general objects from just monocular video. 
In contrast to previous work, our approach does not rely on 3D to 2D correspondences and due to the GPU-based solver architecture it is also much faster than previous approaches. Furthermore, our novel texture term enables tracking of regions with little texture.

\section{Method}
%
%
\label{sec:method}
%
%
%
%
The goal is to estimate the non-rigid deformation of an object from $T$ frames $I^t(x,y)$ with $t \in \{1,...,T\}$.
We assume a static camera and known camera intrinsics.
%
%
Since this problem is in general severely under-constrained, it is assumed that a template triangle mesh of the object to be tracked is given as the matrix $\hat{\vec{V}} \in \mathbb{R}^{N \times 3}$ where each row contains the coordinates of one of the $N$ vertices. 
According to that, $\hat{\vec{V}}_i$ is defined as the $i$th vertex of the template in vector form. 
This notation is also used for the following matrices. 
The edges of the template are given as the mapping~$\mathcal{N}(i)$. 
Given a vertex index $i \in \{1,2,...,N\}$, it returns the set of indices sharing an edge with $\hat{\vec{V}}_i$. 
The $F$ faces of the mesh are represented as the matrix $\vec{F} \in \{1,...,N\}^{F \times 3}$. 
Each row contains the vertex indices of one triangle. 
The UV map is given as the matrix $\vec{U} \in \mathbb{N}^{N \times 2}$. 
Each row contains the UV coordinates for the corresponding vertex. 
The color $\vec{C}_i \in \{0,...,255\}^3$ of vertex $i$ can be computed by a simple lookup in the texture map $I_\mathrm{TM}$ at the position $\vec{U}_i$. 
The color of all vertices is stored in the matrix $\vec{C} \in \{0,...,255\}^{N \times 3}$.
%
%
Furthermore, it is assumed that the geometry at time $t=1$ roughly agrees with the true shape shown in the video so that the gradients of the photometric term can guide the optimization to the correct solution without being trapped into local minima.
The non-rigidly deformed mesh at time $t+1$ is represented as the matrix $\vec{V}^{t+1} \in \mathbb{R}^{N \times 3}$ and contains the updated vertex positions according to the 3D displacement from $t$ to $t+1$. 
%
%
\subsection{Non-rigid Tracking as Energy Minimization}
%
%
Given the template $\hat{\vec{V}}$ and our estimate of the previous frame $\vec{V}^t$, our method sequentially estimates the geometry $\vec{V}^{t+1}$ of the current frame $t+1$. We jointly optimize per-vertex local rotations denoted by $\vec{\Phi}^{t+1}$ and vertex locations $\vec{V}^{t+1}$.
Specifically, for each time step the deformation estimation is formulated as the non-linear optimization problem
\begin{equation}
(\vec{V}^{t+1},\vec{\Phi}^{t+1}) = \argminA_{\vec{V}, \vec{\Phi} \in \mathbb{R}^{N \times 3}} E \left( \vec{V}, \vec{\Phi} \right)\text{,}
\end{equation}
with
\begin{align}
\label{eq:nrenergy}
\begin{split}
E \left( \vec{V}, \vec{\Phi} \right) 
&= \lambda_\mathrm{Photo} 		 E_\mathrm{Photo} 		\left(\vec{V} \right)
+ \lambda_\mathrm{Smooth}   E_\mathrm{Smooth}		\left(\vec{V}	\right)
+ \lambda_\mathrm{Edge} 	 	 E_\mathrm{Edge}		\left(\vec{V}	\right)\\
&+ \lambda_\mathrm{Arap}  E_\mathrm{Arap}	\left(\vec{V}, \vec{\Phi}\right)
+ \lambda_\mathrm{Vel}     E_\mathrm{Vel}		\left(\vec{V}	\right)
+ \lambda_\mathrm{Acc} E_\mathrm{Acc}	\left(\vec{V}	\right)\text{.}
\end{split} 
\end{align}
$\lambda_{Photo}$, $\lambda_{Smooth}$, $\lambda_{Edge}$, $\lambda_{Vel}$, $\lambda_{Acc}$, $\lambda_{Arap}$ are hyperparameters set before the optimization starts and afterwards they are kept constant.
$E \left( \vec{V}, \vec{\Phi} \right)$ combines different cost terms ensuring that the mesh deformations agree with the motion in the video. 
The resulting non-linear least squares optimization problem is solved with the GPU-based Gauss-Newton solver based on the work of Zollhoefer et al.~\cite{zol014} where we adapted the Jacobian and residual implementation to our energy formulation.
The high efficiency is obtained by exploiting the sparse structure of the system of normal equations.
For more details we refer the reader to the approach of Zollhoefer et al.~\cite{zol014}.
Now, we will explain the terms in more detail.
%
%
\par
\textbf{Photometric Alignment.} 
The photometric term
\begin{align}
\begin{aligned}
E_\mathrm{Photo}(\vec{V}) = \sum_{i=1}^{N} 
		  	\euclidean{ \sigma \left( I^{t+1} \ast \vec{G}_w \left( \Pi \left( \vec{V}_i \right) \right) - \vec{C}_{i}  \right) } &^2
\end{aligned}
\end{align}
densely measures the re-projection error. $\euclidean{ \cdot }$ is the Euclidean norm, $\ast$ is the convolution operator and $\vec{G}_w$ is a Gaussian kernel with standard deviation $w$. 
We use Gaussian smoothing on the input frame for more stable and longer range gradients.
$\Pi(\vec{V}_i) = (\frac{u}{w}, \frac{v}{w})^\top$ with $(u,v,w)^\top = \vec{I}\vec{V}_i$ projects the vertex $\vec{V}_i $ on the image plane and $ I^{t+1} \ast \vec{G}_w $ returns the RGB color vector of the smoothed frame at position $\Pi \left( \vec{V}_i \right)$ which is compared against the pre-computed and constant vertex color $\vec{C}_i$.
Here, $\vec{I} \in \mathbb{R}^{3 \times 3} $ is the intrinsic camera matrix.
$\sigma$ is a robust pruning function for wrong correspondences with respect to color similarity. 
More specifically, we discard errors above a certain threshold because in most cases they are due to occlusions.
%
%
%
\par
\textbf{Spatial Smoothness.} 
Without regularization, estimating 3D geometry from a single image is an ill-posed problem. 
Therefore, we introduce several spatial and temporal regularizers to make the problem well-posed and to propagate 3D deformations into areas where information for data terms is missing, e.g., poorly textured or occluded regions.
The first prior
\begin{equation}\label{eq:smooth}
E_\mathrm{Smooth} \left( \vec{V} \right) =  
\sum_{i=1}^{N} 
	\sum_{j \in \mathcal{N}(i)} 
		\euclidean{ \left( \vec{V}_i -  \vec{V}_j \right)-
		( \hat{\vec{V}}_i - \hat{\vec{V}}_j )}^2
\end{equation}
ensures that if a vertex $\vec{V}_i$ changes its position, its neighbors $\vec{V}_j$ with $ j \in \mathcal{N}(i)$ are deformed such that the overall shape is still spatially smooth compared to the template mesh $\hat{\vec{V}}$. 
%
%
In addition, the prior
\begin{equation}
E_\mathrm{Edge} \left( \vec{V} \right) =  
\sum_{i=1}^{N} 
	\sum_{j \in \mathcal{N}(i)}
		\left(\euclidean{ \vec{V}_i - \vec{V}_j } - 
		\lVert \hat{\vec{V}}_i - \hat{\vec{V}}_j \rVert 
	\right)^2
\end{equation}
ensures isometric deformations which means that the edge length with respect to the template is preserved.
In contrast to $E_\mathrm{Smooth}$, this prior is rotation invariant.
%
%
Finally, the as-rigid-as-possible (ARAP) prior~\cite{sor007}
\begin{equation}\label{eq:arap}
E_\mathrm{Arap} \left( \vec{V}, \vec{\Phi} \right) =  
\sum_{i=1}^{N} 
\sum_{j \in \mathcal{N}(i)} 
\euclidean{ \left( \vec{V}_i -  \vec{V}_j \right)-
	\mathcal{R}(\vec{\Phi}_i)( \hat{\vec{V}}_i - \hat{\vec{V}}_j )}^2 \text{,}
\end{equation}
allows local rotations for each of the mesh vertices as long as the relative position with respect to their neighborhood remains the same.
Each row of the matrix $\vec{\Phi} \in \mathbb{R}^{N \times 3}$ contains the per-vertex Euler angles which encode a local rotation around $\vec{V}_i$.
$\mathcal{R}(\vec{\Phi}_i)$ converts them into a rotation matrix.
\par
We choose a combination of spatial regularizers to ensure that our method can track different types of non-rigid deformations equally well. 
For example, $E_\mathrm{Smooth}$ is usually sufficient to track facial expressions without large head rotations. 
But tracking rotating objects can only be achieved with rotational invariant regularizers ($E_\mathrm{Edge}$, $E_\mathrm{Arap}$).
In contrast to Yu et al.~\cite{yu015}, we adopt the Euclidean norm in Eq.~\ref{eq:smooth} and Eq.~\ref{eq:arap} instead of the Huber loss because it led to visually better results.
%
%
%
\par
\textbf{Temporal Smoothness.} 
To enforce temporally smooth reconstructions, we propose two additional priors.
The first one is defined as
\begin{equation}
E_\mathrm{Velocity} \left( \vec{V} \right) =  
	\sum_{i=1}^{N} 
		\euclidean{ \vec{V}_i - \vec{V}_i^t }^2 
\end{equation}
and ensures that the displacement of vertices between $t$ and $t+1$ is small.
%
%
Second, the prior
\begin{equation}
E_\mathrm{Acc} \left( \vec{V} \right) =
	\sum_{i=1}^{N}
		\euclidean{ \left( \vec{V}_i -\vec{V}_i^t \right) -
		\left( \vec{V}_i^t -\vec{V}_i^{t-1} \right) }^2
\end{equation}
penalizes large deviations of the current velocity direction from the previous one.
%
%
\subsection{Non-rigid Tracking of Woven Fabrics}
\label{sec:wovenfabrics}
\begin{figure*}[t]
	\centering
	\includegraphics[width=\textwidth]{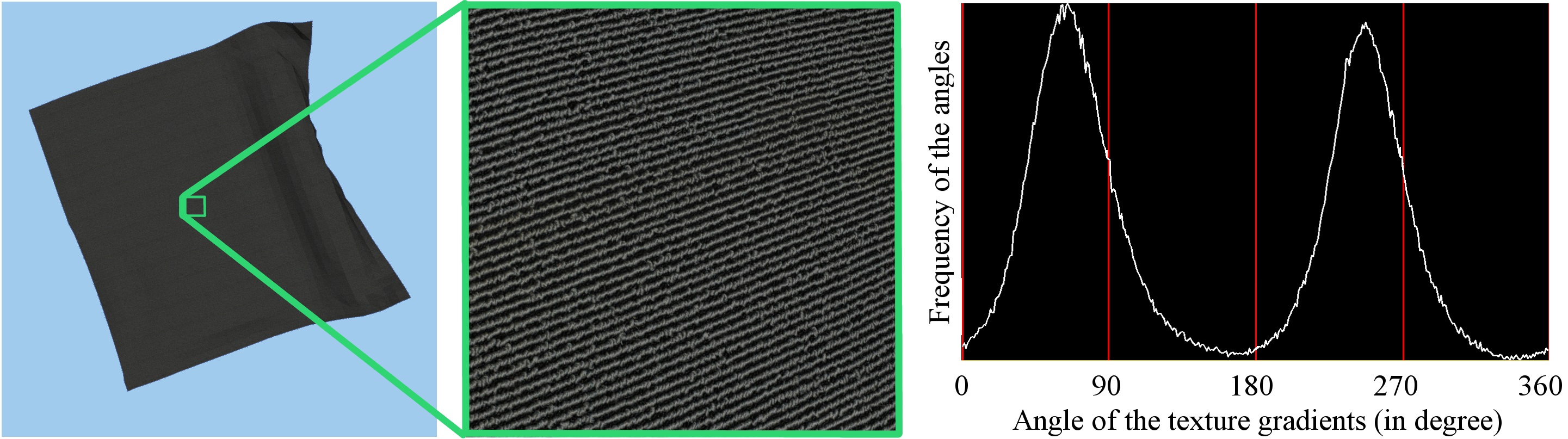}
	\caption
	{
		Histogram of oriented gradients of woven fabrics. 
		Left. 
		The neighborhood region with the center pixel in the middle. 
		Right. 
		The corresponding histogram.
	}
	\label{fig:hogstripes}
\end{figure*}

%
%
Tracking of uniformly colored fabrics is usually challenging for classical color-based terms due to the lack of color features.
To overcome this limitation, we inspected the structure of different garments and found that most of them show line-like micro-structures due to the manufacturing process of the woven threads, see Fig.~\ref{fig:hogstripes} left.
Those can be recorded with recent high resolution cameras such that reconstruction algorithms can make use of those patterns. 
To this end, we propose a novel texture term to refine the estimation of non-rigid motions for the case of woven fabrics.
It can be combined with the terms in Eq.~\ref{eq:nrenergy}.
Now, we will explain our novel data term in more detail. 
%
%
\par
\textbf{Histogram of Oriented Gradient (HOG).} 
Based on HOG~\cite{dal005} we compute for each pixel $(i,j)$ of an image the corresponding histogram $\vec{h}_{i,j} \in \mathbb{R}^b$ where $b$ is the number of bins that count the total number of gradient angles present in the neighborhood of pixel $(i,j)$.
To be more robust with respect to outliers and noise we count the number of gradients per angular bin irrespective of the gradient magnitude and only if the magnitude is higher than a certain threshold.
Compared to pure image gradients, HOG is less sensitive to noise. 
Especially for woven fabrics, image gradients are very localized since changing the position in the image can lead to large differences in the gradient directions due to the high frequency of the image content.
HOG instead averages over a certain window so that outliers are discarded.
%
%
\par
\textbf{Directions of a Texture.} 
Applying HOG to pictures of fabrics reveals their special characteristics caused by the line like patterns (see Fig.~\ref{fig:hogstripes}).
There are two dominant texture gradient angles $\alpha$ and $\beta=((\alpha + 180) \mod 360)$ perpendicular to the lines. 
So $\alpha$ provides the most characteristic information of the pattern in the image at $(i,j)$ and can be computed as the angle whose bin has the highest frequency in $\vec{h}_{i,j}$.
$\alpha$ is then converted to its normalized 2D direction, also called \textit{dominant frame gradient} (DFG), which is stored in the two-valued image $I_{\mathrm{Dir}}(i,j)$.
To detect image regions that do not contain line patterns, we set $I_{\mathrm{Dir}}(i,j)=(0,0)^\top$ if the highest frequency is below a certain threshold.
%
%
\par
\begin{figure*}[t]
	\centering
	\includegraphics[width=\textwidth]{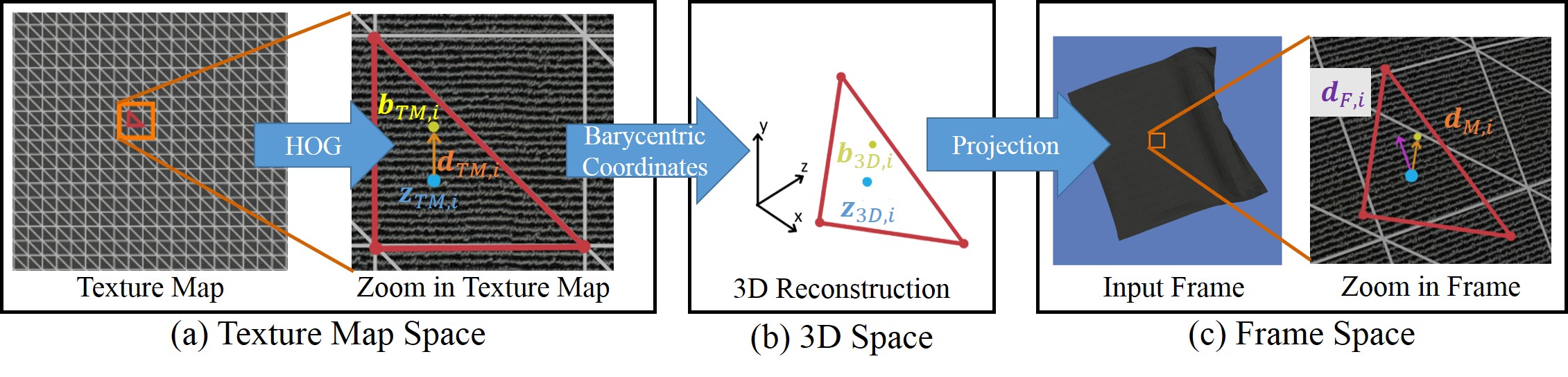}
	\caption
	{
		Overview of the proposed texture term.
	}
	\label{fig:texturetermoverview}
\end{figure*}

\textbf{Texture-based Constraint.} 
Our novel texture term
\begin{equation}
\begin{split}
E_\mathrm{Tex} \left( \vec{V} \right) =  
\sum_{i=1}^{F} 
\euclidean{\rho
	\left(
	\vec{d}_{\mathrm{M},i}
	,
	\vec{d}_{\mathrm{F},i}
	\right)
} ^2
\end{split}
\end{equation}
ensures now that for all triangles the projected DFG $\vec{d}_{\mathrm{M},i}$ parametrized on the object surface agrees with the frame's DFG $\vec{d}_{\mathrm{F},i}$ at the location of the projected triangle center.
An overview is shown in Fig.~\ref{fig:texturetermoverview}.
More precisely, by averaging $\vec{U}_k,\vec{U}_m,\vec{U}_l$ one can compute the pixel position $\vec{z}_{\mathrm{TM},i} \in \mathbb{R}^2$ of the center point of the triangle $\vec{F}_i = (k,m,l)$ in the texture map.
Now, the neighborhood region for HOG around $\vec{z}_{\mathrm{TM},i}$ is defined as the 2D bounding box of the triangle. 
The HOG descriptor for $\vec{z}_{\mathrm{TM},i}$ can be computed and by applying the concept explained in the previous paragraph one obtains the DFG $\vec{d}_{\mathrm{TM},i}$ (see Fig~\ref{fig:texturetermoverview} (a)).
Next, we define $\vec{b}_{\mathrm{TM},i} =\vec{z}_{\mathrm{TM},i} + \vec{d}_{\mathrm{TM},i}$ and express it as a linear combination of the triangles' UV coordinates leading to the \textit{barycentric coordinates} $\vec{B}_{i,1},\vec{B}_{i,2},\vec{B}_{i,3}$ of the face $\vec{F}_i$. 
They form together with the other triangles the barycentric coordinates matrix $\vec{B} \in \mathbb{R}^{F \times 3}$.
Each row represents the texture map's DFG for the respective triangle of the mesh in an implicit form.
Since $\vec{b}_{\mathrm{TM},i}$ can be represented as a linear combination, one can compute the corresponding 3D point $\vec{b}_{\mathrm{3D},i} = \vec{B}_{i,1} \vec{V}_k + \vec{B}_{i,2}\vec{V}_m + \vec{B}_{i,3}\vec{V}_l $ as well as the triangle center $\vec{z}_{\mathrm{3D},i} \in \mathbb{R}^3$ in 3D (see Fig~\ref{fig:texturetermoverview} (b)).
The barycentric coordinates remain constant, so that $\vec{b}_{\mathrm{3D},i}$ and $\vec{z}_{\mathrm{3D},i}$ only depend on the mesh vertices $\vec{V}_k$, $\vec{V}_m$ and $\vec{V}_l$. 
One can then project the DFG of the mesh $\vec{d}_{\mathrm{M},i} = \Pi(\vec{b}_{\mathrm{3D},i})- \Pi(\vec{z}_{\mathrm{3D},i})$ into the frame and compare it against the DFG $\vec{d}_{\mathrm{F},i}$ of the frame $t+1$ at the location of $\Pi(\vec{z}_{\mathrm{3D},i})$ which can be retrieved by an image lookup in $I_{\mathrm{Dir}}^{t+1}$ (see Fig~\ref{fig:texturetermoverview} (c)).
$\rho\left( \vec{x},\vec{y} \right)$ computes the minimum of the differences between $\vec{x},\vec{y}$ and $\vec{x},-\vec{y}$ iff both $\vec{x}$ and $\vec{y}$ are non-zero vectors (otherwise we are not in an area with line patterns) and the directions are similar up to a certain threshold to be more robust with respect to occlusions and noise. 
As mentioned above, there are two DFGs in the frame for the case of line patterns. 
We assume the initialization is close to the ground truth and choose the minimum of the two possible directions. 
\section{Results}
\label{sec:results}
%
%
All experiments were performed on a PC with an NVIDIA GeForce GTX 1080Ti and an Intel Core i7.
In contrast to related methods~\cite{yu015}, we achieve interactive frame rates using the energy proposed in Eq.~\ref{eq:nrenergy}. 
%
%
\subsection{Qualitative and Quantitative Results}
\label{sec:synrig}
%
%
Now, we evaluate NRST on datasets for general objects like faces where we disable $E_\mathrm{Tex}$.
After that, we compare our approach against another monocular method.
Finally, we evaluate our proposed texture term on two new scenes showing line-like fabric structures, perform an ablation study and demonstrate interesting applications.
More results can be found in the supplemental video.
%
%
\begin{figure*}[t]
	\centering
	\includegraphics[width=\textwidth]{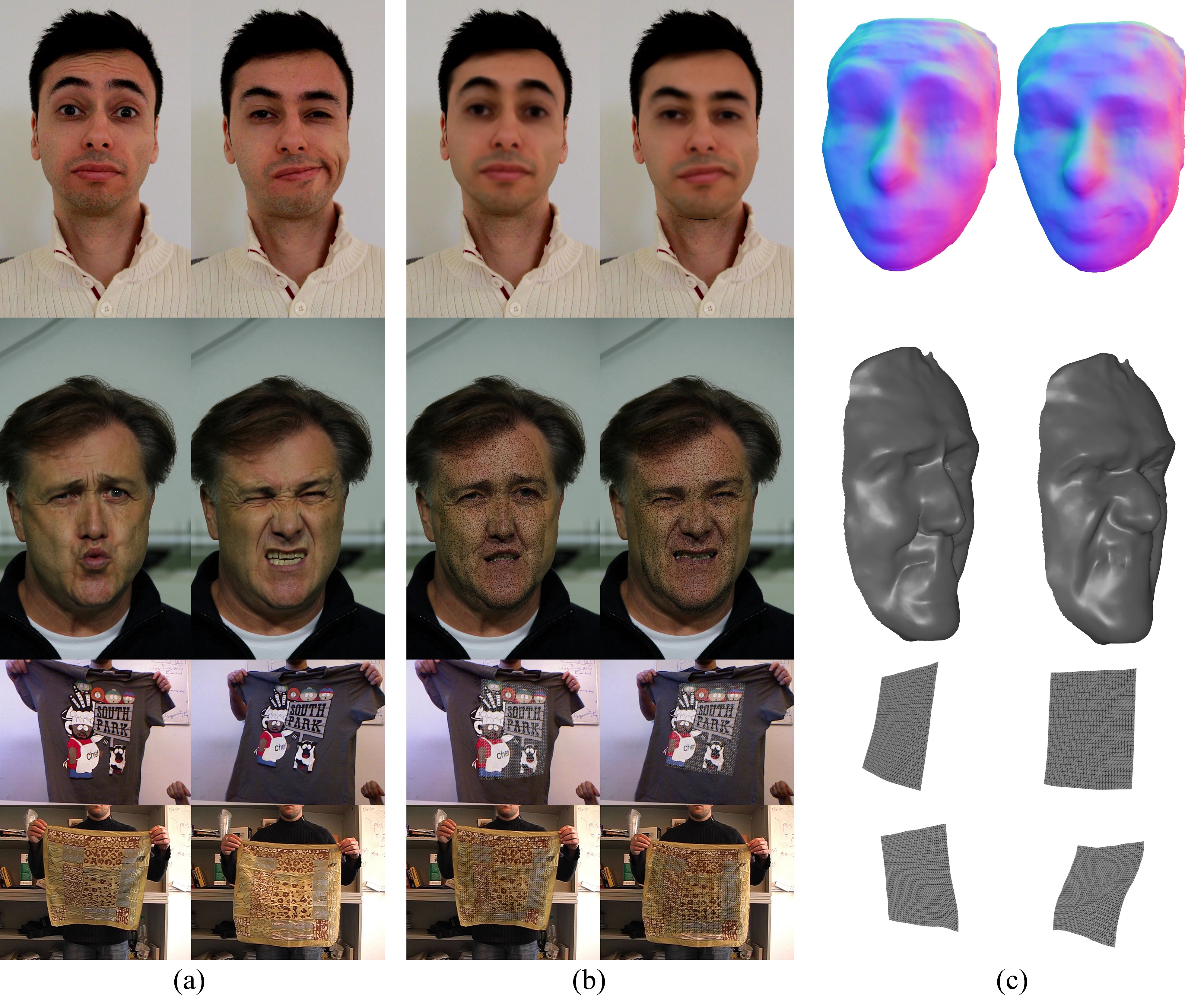}
	\caption
	{
		Reconstructions of existing datasets~\cite{yu015,valg012,var012,sal008}.
		Input frames (a). 
		Textured reconstructions overlayed on the input (b). 
		Deformed geometries obtained by our method (rendered from a different view point) (c).
	}
	\label{fig:QualitativeGeneral}
\end{figure*}
\par
\textbf{Qualitative Evaluation for General Objects.}
%
%
In Fig.~\ref{fig:QualitativeGeneral} we show frames from our monocular reconstruction results.
%
%
We tested our approach on two face sequences~\cite{yu015,valg012} where templates are provided. 
Note that NRST precisely reconstructs facial expressions.
The 2D overlay (second column) matches the input and also in 3D (third column) our results look realistic.
%
%
Furthermore, we evaluated on the datasets of Varol et al.~\cite{var012} and Salzmann et al.~\cite{sal008} showing fast movements of a T-shirt and a waving towel.
Again for most parts of the surface the reconstructions look accurate in 2D since they overlap well with the input and they are also plausible in 3D.
This validates that our approach can deal with the challenging problem of estimating 3D deformations from a monocular video for general kinds of objects.
%
%
\par
\begin{figure*}
	\centering
	\includegraphics[width= 0.6\textwidth]{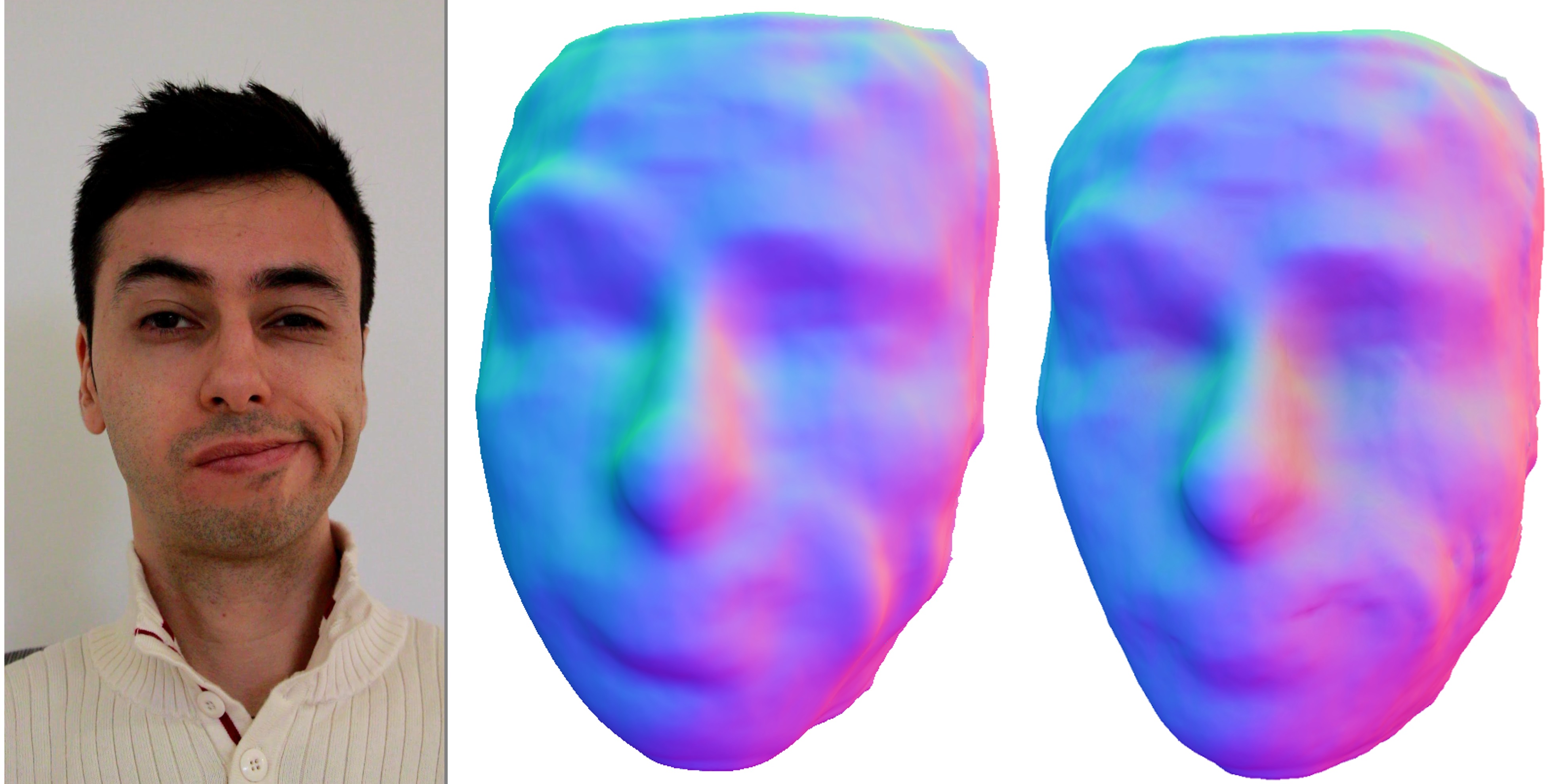}
	\caption
	{
		Comparison of NRST's reconstruction (right) and the one of Yu et al.~\cite{yu015} (middle). 
		It becomes obvious that both capture the facial expression shown in the input frame (left), but the proposed approach is significantly faster than the one of Yu et al. due to our data-parallel GPU implementation.
	}
	\label{fig:compyu}
\end{figure*}

\textbf{Comparison to Yu et al.~\cite{yu015}.} 
Fig.~\ref{fig:compyu} shows a qualitative comparison between our method and the one of Yu et al.~\cite{yu015}.
It becomes obvious that both capture the facial expression, but the proposed approach is faster than the one of Yu et al. due to our data-parallel GPU implementation. 
In particular, on their sequence our method runs at 15fps whereas their approach takes several seconds per frame.
More sidy-by-side comparisons on this sequence can be found in the supplemental video.
\begin{figure*}
	\centering
	\includegraphics[width=\textwidth]{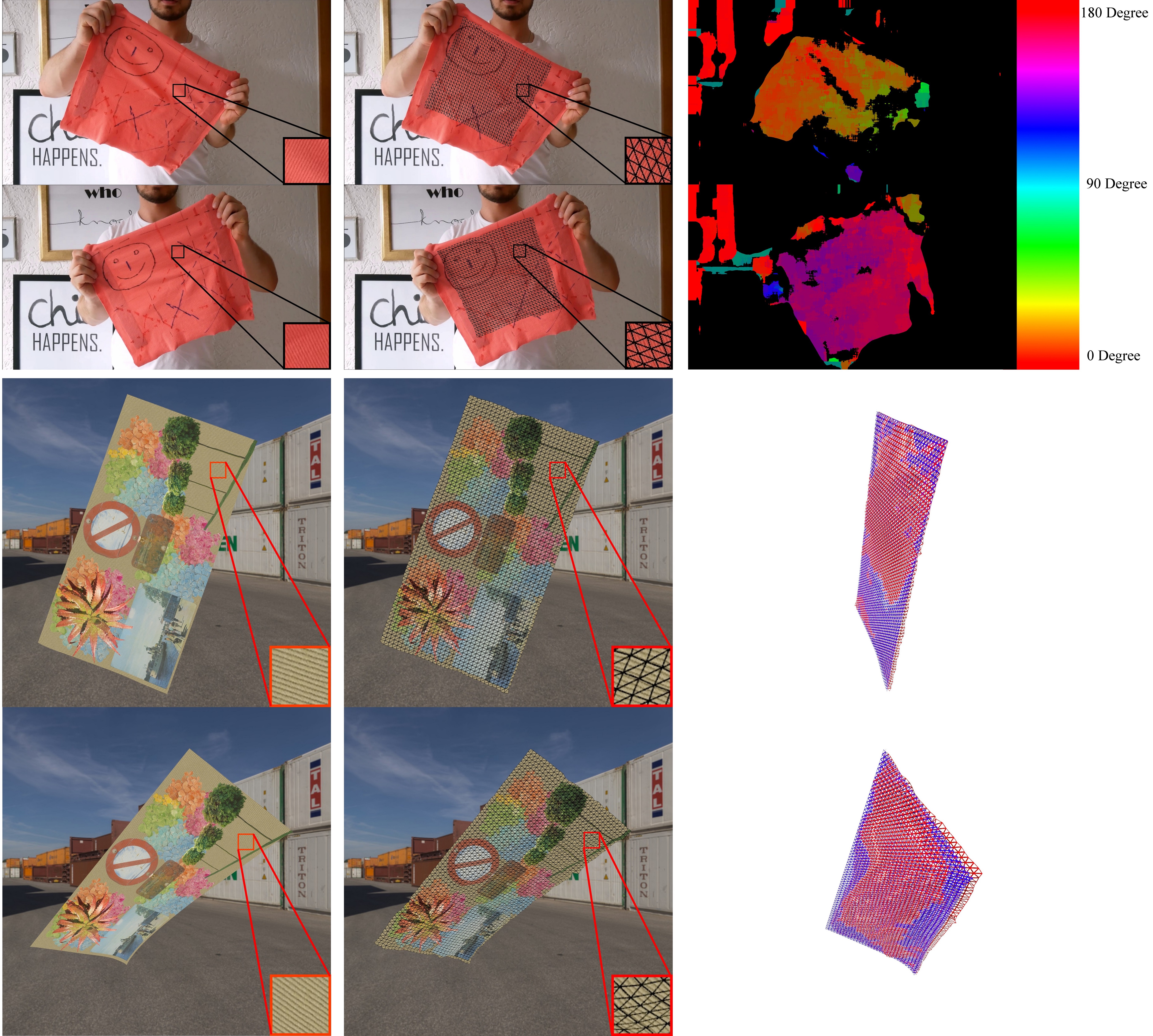}
	\caption
	{
		Reconstruction of line patterns. 
		Top from left to right. 
		Input frames. 
		Textured reconstructions overlayed on the frames. 
		Color coded visualization of the estimated dominant frame angles. 
		Regions where no line pattern was detected are visualized in black.
		Bottom from left to right. 
		Input frames. 
		Textured reconstructions overlayed on the frames. 
		Ground truth geometries (blue) and our reconstructions (red).
	}
	\label{fig:QualitativeTextureTerm}
\end{figure*}
%
%
\par
\textbf{Qualitative Evaluation for Fabrics.}
%
%
The top row of Fig.~\ref{fig:QualitativeTextureTerm} shows frames (resolution $1920 \times 1080$) of a moving piece of cloth that has the typical line patterns. 
Although the object is sparsely textured, our approach is able to recover the deformations due to the texture term, which accurately tracks the DFG of the line pattern.
As demonstrated in the last column, the estimated angles for the frames are correct and therefore give a reliable information cue exploited by $E_\mathrm{Tex}$.
%
%
For quantitative evaluation, we created a synthetic scene that is modeled and animated in a modeling software showing a carpet that has the characteristic line pattern but is also partially textured (see bottom row of Fig.~\ref{fig:QualitativeTextureTerm}).
We rendered the scene at a resolution of $1500 \times 1500$.
$E_\mathrm{Tex}$ helps in the less textured regions where $E_\mathrm{Photo}$ would fail. 
The last column shows how close our reconstruction (red) is with respect to ground truth (blue).
%
%
\par
\begin{figure*}[t]
	\centering
	\includegraphics[width=\textwidth]{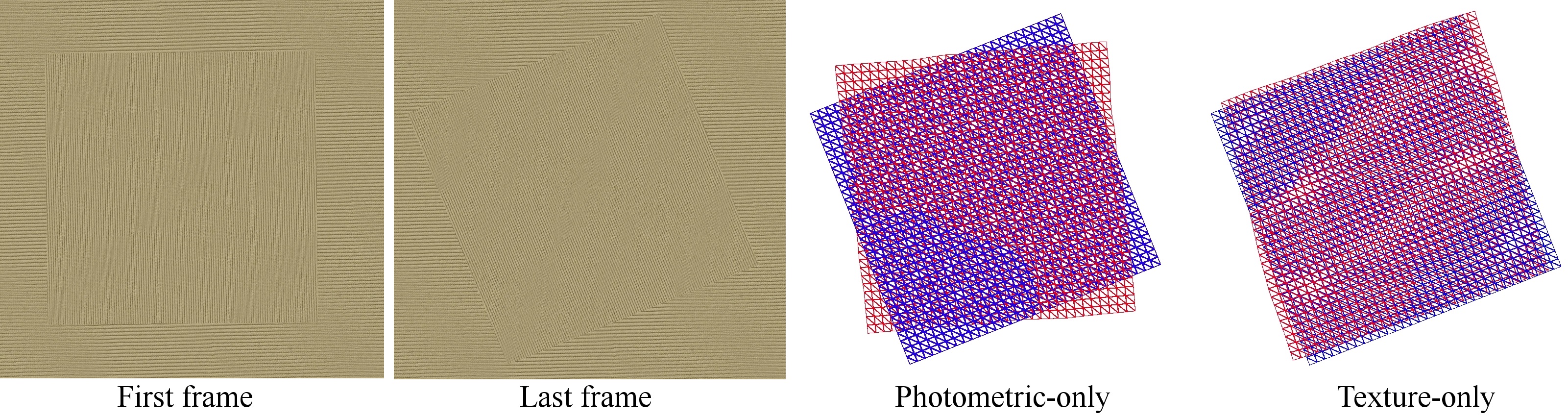}
	\caption
	{
		Rotating object sequence. 
		From left to right. 
		First and last frame. 
		Note that the object and the background have the same color. 
		The reconstructions (red) of the last frame with either $E_\mathrm{Photo}$ or $E_\mathrm{Tex}$ overlayed on the ground truth geometry (blue). 
		Note that $E_\mathrm{Tex}$ can recover the rotation in contrast to $E_\mathrm{Photo}$.
	}
	\label{fig:ablation}
\end{figure*}

\textbf{Ablation Analysis.} 
Apart from the proposed texture term, our energy formulation is similar to the one of Yu et al.~\cite{yu015}.
To validate that $E_\mathrm{Tex}$ improves the reconstruction over a photometric-only formulation, we perform an ablation study.
We measured the averaged per-vertex Euclidean distance between the ground truth mesh and our reconstructions.
For the waving towel shown in Fig.~\ref{fig:QualitativeTextureTerm} bottom, we obtained an error of 26.8mm without $E_\mathrm{Tex}$ and 25.5mm if we also use our proposed texture term leading to an improvement of 4.8\%. 
The diagonal of the 3D bounding box of the towel is 3162mm.
For the rotation sequence (resolution $800 \times 800$) shown in Fig.~\ref{fig:ablation} the color variation is very limited since background and object have the same color.
In contrast to $E_\mathrm{Photo}$ alone, $E_\mathrm{Tex}$ can rotate the object leading to an error of 4.1mm for the texture-only case and 6.7mm for the photometric-only setting.
So $E_\mathrm{Tex}$ improves over $E_\mathrm{Photo}$ by 38.8\%
%
%
\subsection{Applications}
\begin{figure*}[t]
	\centering
 	\includegraphics[width=\textwidth]{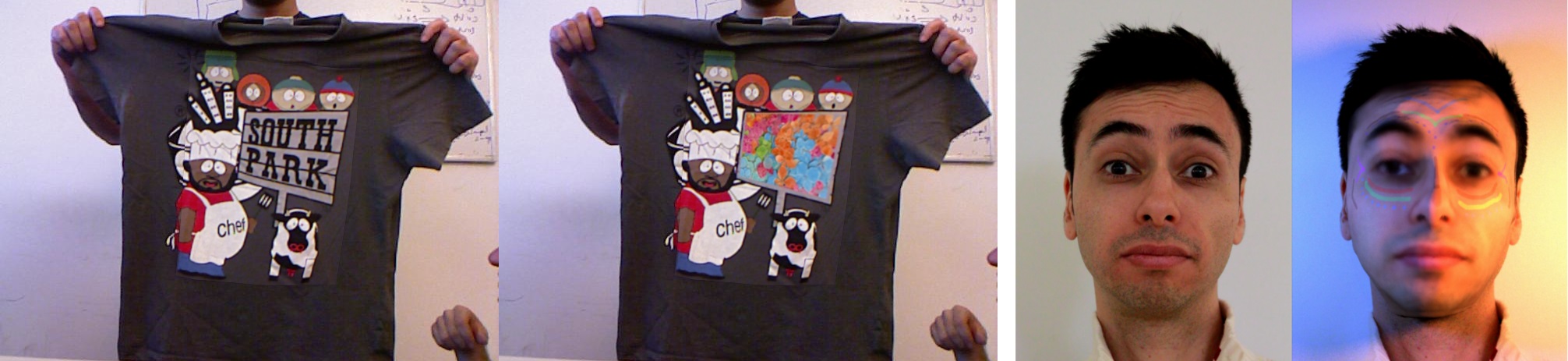}
	\caption
	{
		Applications. 
		Left. 
		Re-textured shirt. 
		Right. 
		Re-textured and re-lighted face.
	}
	\label{fig:reprojection}
\end{figure*}

Our method enables several applications such as free view point rendering or re-texturing on general deformable objects or for virtual face make-up (see Fig.~\ref{fig:reprojection}). 
Since our approach estimates the deforming geometry, one can even change the scene lighting for the foreground such that the shading remains realistic. 
%
%
\subsection{Limitations}
By the nature of the challenging task of monocular tracking of non-rigid deformations, our method has some limitations which open up directions for future work.
%
%
Although, our proposed texture term uses more of the information contained in the video than a photometric-only formulation, there are still image cues that can improve the reconstruction like shading and the object contour as demonstrated by previous work~\cite{liu016,xu017}. 
So, one could combine them in a unified framework.
%
%
To increase robustness, the deformations could be jointly estimated over a temporal sliding window as proposed by Xu et al.~\cite{xu017} and an embedded graph~\cite{sum007} could lead to improved stability by reducing the number of unknowns.

	\section{Conclusion}
\label{sec:conclusion}
We presented an optimization-based analysis-by-synthesis method that solves the challenging task of estimating non-rigid motion, given a single RGB video and a template. 
Our method tracks non-trivial deformations of a broad class of shapes, ranging from faces to deforming fabric. 
Further, we introduce specific solutions tailored to capture woven fabrics, even if they lack clear color variations.
Our method runs at interactive frame rates due to the GPU-based solver that can efficiently solve the non-linear least squares optimization problem.
Our evaluation shows that the reconstructions are accurate in 2D and 3D which enables several applications such as re-texturing.
	\paragraph{Acknowledgments.}{This work was funded by the ERC Consolidator Grant 4DRepLy (770784).}
	%
	%
	%
	%
	\bibliographystyle{splncs04}
	\bibliography{egbib}
\end{document}